\documentclass[conference]{IEEEtran}
\usepackage{cite}
\usepackage{amsmath,amssymb,amsfonts}
\usepackage{algorithmic}
\usepackage{graphicx}
\graphicspath{ {./figs/} }
\usepackage{multirow}
\usepackage{subfigure}
\usepackage{cuted}

\usepackage{array}
\usepackage{tabularx}
\usepackage{hhline}
\usepackage{xspace}
\usepackage{upgreek}
\usepackage{makecell}

\newcounter{itemlistc}

\newenvironment{itemlist}
    {   \setcounter{itemlistc}{0}
    \begin{list}{$\bullet$}
        {\usecounter{itemlistc}
        \setlength{\parsep}{0pt}
        \setlength{\topsep}{3pt}
        \setlength{\itemsep}{0pt}}
        }{ \end{list} }

\def\BibTeX{{\rm B\kern-.05em{\sc i\kern-.025em b}\kern-.08em
    T\kern-.1667em\lower.7ex\hbox{E}\kern-.125emX}}

\IEEEoverridecommandlockouts    
\begin{document}

\title{Enhanced DeepONet for Modeling Partial Differential Operators Considering Multiple Input Functions
\thanks{This paper was performed in the summer 2021 and the
  manuscript was submitted for review in Sept 2021.}}

\author{\IEEEauthorblockN{Lesley Tan$^1$ and Liang Chen$^2$}
\IEEEauthorblockA{$^1$Phillips Academy, Andover, MA 01810}
\IEEEauthorblockA{$^2$Department of Electrical and Computer Engineering, University of California, Riverside, CA 92521}
\IEEEauthorblockA{ltan22@andover.edu, liangch@ucr.edu}
}

\maketitle
\begin{abstract}
  Machine learning, especially deep learning is gaining much attention
  due to the breakthrough performance in various cognitive
  applications.  Recently, neural networks (NN) have been intensively
  explored to model partial differential equations as NN can be viewed
  as universal approximators for nonlinear functions. A deep network
  operator (DeepONet) architecture was proposed to model the general
  non-linear continuous operators for partial differential equations
  (PDE) due to its better generalization capabilities than existing
  mainstream deep neural network architectures. However, existing
  DeepONet can only accept one input function, which limits its
  application.  In this work, we explore the DeepONet architecture to
  extend it to accept two or more input functions. We propose new
  Enhanced DeepONet or {\it EDeepONet} high-level neural network
  structure, in which two input functions are represented by two
  branch DNN sub-networks, which are then connected with output truck
  network via inner product to generate the output of the whole neural
  network.  The proposed {\it EDeepONet} structure can be easily
  extended to deal with multiple input functions.  Our numerical
  results on modeling two partial differential equation examples shows
  that the proposed enhanced DeepONet is about {\bf 7X-17X} or about
  {\bf one order of magnitude} more accurate than the fully connected
  neural network and is about {\bf 2X-3X} more accurate than a simple
  extended DeepONet for both training and test.
\end{abstract}


\section{Introduction}

Deep neural networks (DNN) have propelled an evolution in machine learning fields and redefined many existing applications with new human-level AI capabilities. DNNs such as convolution neural networks (CNN) have recently been applied to many cognitive applications such as visual object recognition, object detection, speech recognition, natural language understanding, etc. due to dramatic accuracy improvements in those
tasks~\cite{LeCun:2015dt,Goodfellow:Book'2016}. Despite the remarkable success in those areas, deep learning has not yet been widely used in the field of scientific computing.  

One particular area of such scientific computing is to model, predict and interpret the physical phenomena via mathematical models. Physics laws such as the Newton's law for motion, forces and gravity, thermodynamics for energy and process of work were modeled via existing mathematical tools. While existing mathematical tools such as calculus have served science well, new fundamental development is slow and has not established any governing principles in some fields such as emerging field of social dynamics. Classical calculus and partial differential equations can be efficient for many well established physics laws, but they may be less efficient or even inadequate to represent much more complicated inter-connected systems~\cite{West16}. From scientific discovery prospective, one important problem is how to efficiently  describe dynamic systems (or its operators -- mapping from a space of functions into another space of functions )  implicitly based on a given data set.  

On the other hand, deep neural networks, which can be viewed as universal approximator, provide new possible of alternative modeling paradigm  for those and  more complicated systems without any prior knowledge~\cite{Cybenko:1989vo,Hornik:NN'1989}.   Recently, neural networks (NN) have been intensively explored to predict partial differential equations as nonlinear functions and nonlinear operators~\cite{DeepXDE_arxiv_2019,Lu:NMI'21}. This work is based on previously proposed {\it  universal function and operator approximation theorems}  using neural network method~\cite{Chen:TNN'93,Chen:TNN'95}. But instead of using one or two layers  neural networks, a deep network operator (DeepONet) architecture was proposed to model the general non-linear continuous operators for partial differential equations (PDE) due to its faster  convergence rate and better generalization capabilities than existing mainstream deep neural network architectures. 
 
 However, existing DeepONet can only accept  one input function such as one boundary condition of PDE and it did not consider two or more input functions such as both initial condition and one boundary conditions for PDE general,   which limits its applications for many practical problems. In this work, we explore the DeepONet architecture to extend it to accept two or more input functions.   The key contributions of this paper are as follows.
 
\begin{itemlist}
 \item First, based on the DeepONet structure, we propose two architectures to consider the two input functions for PDE operators. The first method is just to merge the two function inputs into one hybrid input function, which can be viewed as simple extended DeepONet baseline method in addition to the fully connected neural network method. 
 
 \item Second, we propose new Enhanced DeepONet or {\it EDeepONet} structure, in which two input functions are represented by two branch DNN sub-networks, which are then connected with output truck network via inner product to generate the output of the whole neural network.  The proposed {\it EDeepONet} structure can be easily extended to deal with multiple input functions. 
 
 \item  Our numerical results on modeling two partial differential equation examples show that the proposed {\it EDeepONet} is about  $7X$-$17X$ or{\bf  one order of magnitude}  more accurate than the fully connected neural network and is about $2X$-$3X$ more accurate than the simple extended DeepONet for both training and test. Similar to DeepONet, the proposed {\it EDeepONet} also demonstrates faster convergence rate than fully connected neural network, which enables fast training via small number of  scatted data. 
\end{itemlist}
 
 The rest of the paper is organized as follows: Section~\ref{sec:review} will review related works including the universal function and universal operator approximation theorems and the recently proposed DeepONet architecture for nonlinear operator approximation. Section~\ref{sec:proposed_method} presents the proposed neural network architecture called enhanced DeepONet or {\it EDeepONet} architecture for operator approximation. We will show one flat and one hierarchical architectures implementation of the proposed {\it EDeepOnet}.  We also present two simple implementation of operator approximation methods considering two input functions based on existing fully connected neural networks and the existing DeepONet architecture, which will serve as our two baselines for comparison.   Section~\ref{sec:results} presents the numerical results of the proposed {\it EDeepONet} and compare it against  the two baseline architectures. Finally, section~\ref{sec:concl} summarizes the key contribution of this work and discuss some potential future work. 
\section{Review of related works}
\label{sec:review}
\subsection{The universal nonlinear function approximation}
In this section, we briefly review a few important universal approximation  results for functional and operator based on neural networks. 

\begin{figure}[!ht]
    \centering
    \includegraphics[width=0.49\linewidth]{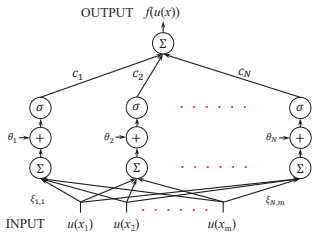}
    \caption{Neural network structure for universal function approximation}
    \label{fig:nn_for_func_approx}
\end{figure}

For a continuous function $f(u)$ defined on $U, u \in U$, where $U$ is a compact set in $C[a,b]$ and $\sigma$ is a bound function like sigmoidal function or other nonlinear function, then $f(u)$ can be approximated by the following expression~\cite{Chen:TNN'93}:
\begin{equation}
\label{eq:func_approx}
    \left|f(u)-\sum_{i=1}^N c_i\sigma\left(\sum_{j=1}^m\xi_{i,j}u(x_j)+\theta_i\right)\right|<\epsilon
\end{equation}
where $\epsilon$ is a given error bound.  $c_i$, $\xi_{i,j}$ and $\theta_i$ are real numbers, $u(x_j)$ value of input function $u(x)$ evaluated at $m$ points $\{x_1, x_2, ..., x_m\}$, which is also called sensors later in~\cite{Lu:NMI'21}. $N$ is the number of neurons in the hidden layer and is a hyper-parameter for this resulting neural network. 
This is also called universal approximation theorem for general linear and nonlinear functions. 
Fig.~\ref{fig:nn_for_func_approx} shows the architecture of the neural network implementation of \eqref{eq:func_approx}. As we can see it basically has one fully connect layer of hidden layer. 

\subsection{The universal nonlinear operator approximation}
Later on, Chen \& Chen~\cite{Chen:TNN'95} further extended their work to consider to approximate dynamic systems described by partial differential equation (PDE), which are called operators. Specifically, $G(u)(y)$ which maps from function $u(x)$ to function $G(y)$, can be approximated by the following expression:
\begin{strip}
\begin{align}
    \left|G(u)(y)-\sum_{k=1}^N\left[\sum_{i=1}^M c_i^{k} g\left(\sum_{j=1}^{m_{k}} \xi_{i,j}^{k} u(x_j)+\theta_i^{k}\right)\right]g\left(\sum_{l=1}^{n} \omega_{k,l} y_l+\zeta_k\right)\right|<\epsilon
\label{eq:oper_approx}
\end{align}
\end{strip}
where $c_i^k$, $\xi_{i,j}^k$, $\theta_i^k$, $\omega_{k,l}$ and $\zeta_k$ are constants, $i=1,\cdots, M, k=1,\cdots, N, l=1,\cdots, N$. $G$ is a nonlinear continuous operator. Again $g$ is a nonlinear activation function like sigmodial or Rectified linear unit (ReLU) functions. Again $N$ and $M$ are hyper-parameters of the resulting neural network. 

Conceptually, equation~\eqref{eq:oper_approx} was derived by applying the universal function approximation theorem of \eqref{eq:func_approx} for function $G(y)$ first and then treat each $c_i$ in \eqref{eq:func_approx} as a function of $G(u)$, i.e. $c_i(G(u))$, which is further approximated by \eqref{eq:func_approx} for each $i$. 

The resulting neural network architecture is shown in Fig.~\ref{fig:nn_for_oper_approx} for \eqref{eq:oper_approx}. As we can see that  it contains multiple fully connected layers. The bottom fully connected layers taking care of the input functions $u(x_i)$ and the upper fully connected layer is responsible for the output variable $y_i$. The two part of networks are connected by element-wise production and summation for total of $N$ intermediate internal neurons.  

\subsection{DeepONet architecture}
\begin{figure}[!ht]
    \centering
    \includegraphics[width=0.8\linewidth]{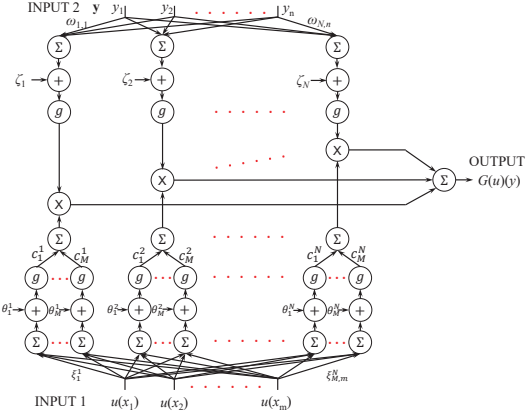}
    \caption{Neural network architecture for universal nonlinear operator approximation}
    \label{fig:nn_for_oper_approx}
\end{figure}

\begin{figure}[!ht]
    \centering
    \subfigure[]{\includegraphics[width=0.49\linewidth]{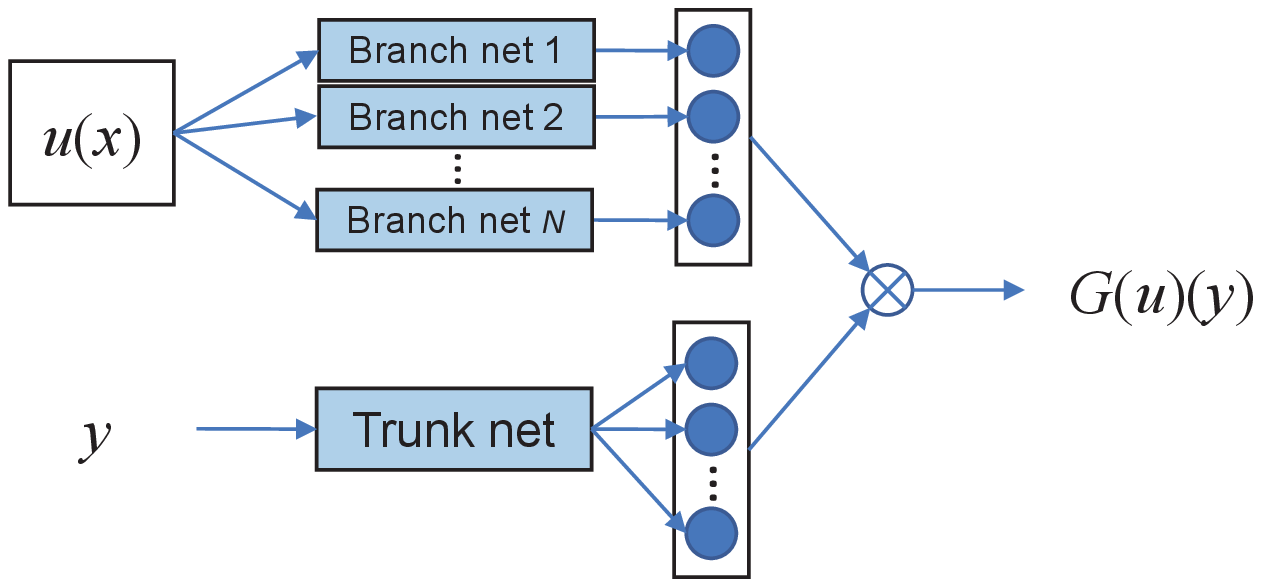}\label{fig:deeponet2}}
    \subfigure[]{\includegraphics[width=0.49\linewidth]{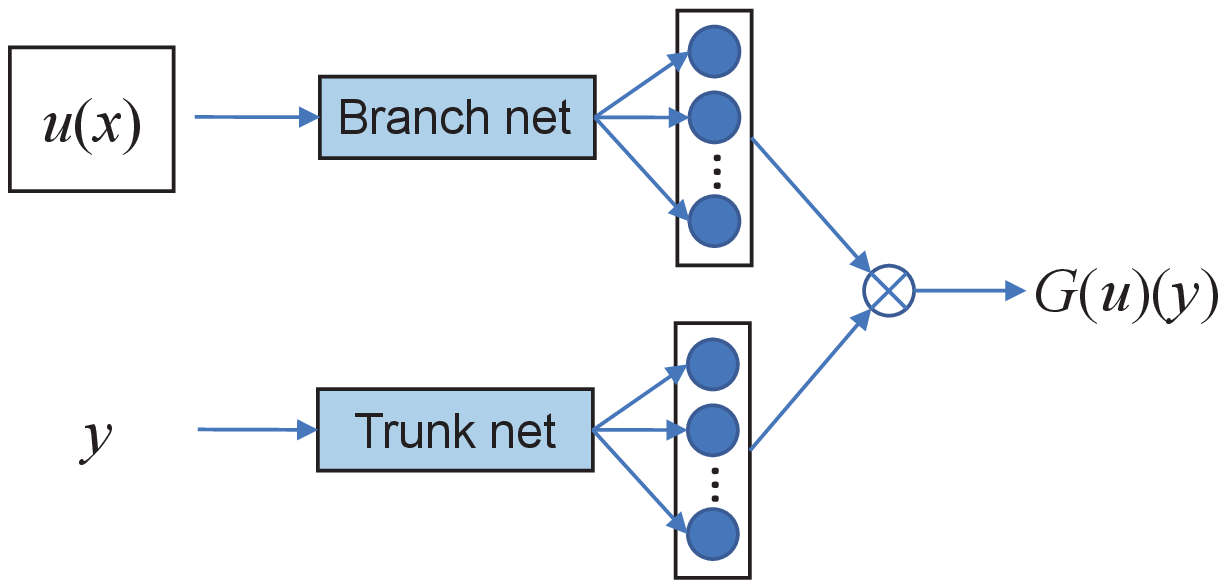}\label{fig:deeponet1}}
    \caption{(a)Stacked DeepONet. (b)Unstacked DeepONet.}
\end{figure}

\begin{strip}
\begin{align} 
\left|G(u)(y)-\sum_{k=1}^{N} \underbrace{\left[\sum_{i=1}^{M} c_{i}^{k} g\left(\sum_{j=1}^{m_k} \xi_{i,j}^{k} u\left(x_{j}\right)+\theta_{i}^{k}\right)\right]}_{\text{branch}} \underbrace{g\left(\sum_{l=1}^n w_{k,l}y_l+\zeta_{k}\right)}_{\operatorname{trunk}}\right|<\epsilon
\label{eq:deeponet_unstacked}
\end{align}
\end{strip}
Inspired by the shallow neural network structure shown in Fig.~\ref{fig:nn_for_oper_approx}, 
Recently, Lu and Karniadakis {\it et al} proposed {\it DeepONet} architecture~\cite{Lu:NMI'21}. Basically they name the network structure associated with the input function as the {\it branch} network and the network structure associated with the output variable $y$ as {\it truck} network. As a result, the \eqref{eq:oper_approx} can be rewritten into the format \eqref{eq:deeponet_unstacked}, in which the sub-expressions associated with {\it branch} and {\it truck} are explicitly marked.  Note that branch net contains several sub fully connected layers. The high level architecture of  stacked DeepONet is shown in Fig.~\ref{fig:deeponet2}.

After that, Lu and Karniadakis {\it et al}  further proposed unstacked DeepONet~\cite{Lu:NMI'21,Lu:SIAM'21}, called generalized universal approximation theorem for operator, which is expressed as
\begin{equation}
\label{eq:univ_op_approx}
    \left|G(u)(y)-\langle\underbrace{\mathbf{g}\left(u\left(x_{1}\right), u\left(x_{2}\right), \cdots, u\left(x_{m}\right)\right)}_{\text {branch }}, \underbrace{\mathbf{f}(y)}_{\text {trunk }}\rangle\right|<\epsilon
\end{equation}
based on the observation that the {\it branch} and {\it truck} networks can be represented by any deep neural networks structures of the shallow networks as shown in Fig.~\ref{fig:nn_for_oper_approx}. Essentially this network merges several sub branch nets into one branch net, which is shown in Fig.~\ref{fig:deeponet1}. Their numerical results on a few partial differential equation examples show that the stacked DeepONet actually achieves better results in terms of training errors, optimization error and generalization errors~\cite{Lu:NMI'21}.

\section{The Proposed EDeepONet considering multi-input functions}
\label{sec:proposed_method}

In this section,  we first  present the new problem that we try to solve, which is to build NN models for nonlinear operator for partial differential equations considering two or multiple input functions.  

Then we present the new neural network architecture called {\it Enhanced DeepOnet} or {\it EDeepONet} structure,  based on the universal nonlinear operator approximation theorem and the improved DeepONet structure.  

\subsection{Several functions as inputs}
\label{sec:problem}
For partial differential equations, the existing NN based approximators only considers one input function for the PDE such as boundary condition or initial condition etc~\cite{Chen:TNN'93,Chen:TNN'95, Lu:NMI'21}.  However, in practice, we may have to consider multiple functions as inputs such as initial conditions, boundary conditions, and parts of PDE, such as coefficients, excitation etc. As a result, this will limit their applications for many real problems. 
It is very important to build a suitable neural network to handle several functions as input. 

In this work, we try to resolve this issue by extending the DeepONet framework to consider multi-input functions. But given the space limitation, we only show the results for two input functions cases and the proposed method can be trivially extended to multi-input cases. 

\begin{figure}[!ht]
    \centering
    \includegraphics[width=0.8\linewidth]{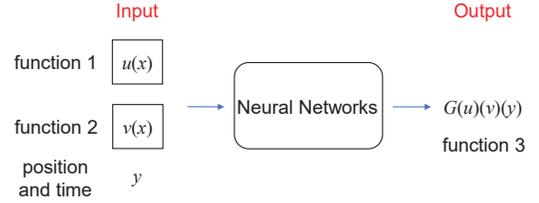}
    \caption{Problem formulation: two functions as input.}
    \label{fig:problem}
\end{figure}

Specifically, the problem can be further illustrated in Fig.~\ref{fig:problem}, which is a mapping from function $u(x)$ and $v(x)$ to function $G(u)(v)(y)$. The inputs are two functions $u(x)$ and $v(x)$, and specific position and time $y$ of output function $G(u)(v)(y)$. 

Before we present our proposed work, we first show how  the new operator modeling problem can be handled by existing fully connected neural networks (FNN) and simple extension of DeepONet. 

\subsection{FNN for two input function operator modeling}
\begin{figure}[!ht]
    \centering
    \includegraphics[width=0.65\linewidth]{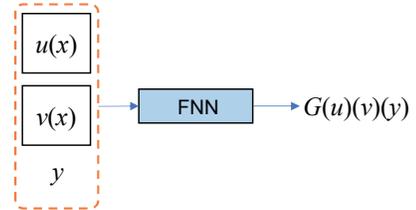}
    \caption{FNN implementation for operator approximation considering two input fucntions}
    \label{fig:fnn}
\end{figure}

The first straightforward method for the problem that is illustrated in Section~\ref{sec:problem} is traditional FNN, which is shown in Fig.~\ref{fig:fnn}.
The idea is to concatenate two input function $u(x)$, $v(x)$ and $y$ into one input vector for the FNN model, after this, we can build a relationship between input and output via the FNN. Mathematically, the FNN can be represented by the expression:
\begin{equation}
    \bigg|G(u)(v)(y)-\text{FNN}(u(x)||v(x)||y)\bigg|<\epsilon
\end{equation}
FNN based method will serve as the first baseline method for the proposed method. 

\subsection{Simple extension of DeepONet for two input function operator modeling}
\begin{figure}[!ht]
    \centering
    \includegraphics[width=0.75\linewidth]{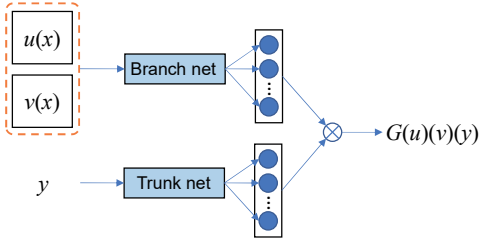}
    \caption{Simple DeepONet extension for operator approximation for two input functions}
    \label{fig:deeponet}
\end{figure}

The second method, is based on existing DeepOnet structure~\cite{Lu:NMI'21,Lu:SIAM'21}. The idea is similar to the FNN model, in which we also concatenate two functions $u(x)$ and $v(x)$ into one input function  and fed them into branch net since the purpose of branch net is to deal with function input. Similarly, the resulting simple two input function DeepONet can be expressed as
\begin{strip}
\begin{align}
    \left|G(u)(v)(y)-\langle\underbrace{\mathbf{g}\left(u\left(x_{1}\right), u\left(x_{2}\right), \cdots, u\left(x_{m}\right),v\left(x_{1}\right), v\left(x_{2}\right), \cdots, v\left(x_{n}\right)\right)}_{\text {branch }}, \underbrace{\mathbf{f}(y)}_{\text {trunk }}\rangle\right|<\epsilon
\end{align}
\end{strip}
The architecture of simple two input function  DeepONet is shown in Fig.~\ref{fig:deeponet}.
However, we show in our numerical result section that such simple extension of DeepONet does not yield the best results. We will use such simple extended DeepONet as our second baseline method.

\subsection{Enhanced DeepONet for two input function operator modeling}
In this section, we present the proposed Enhanced DeepONet structure, called {\it EDeepONet},  to consider the two or more input functions. Our method is still based on the high-level network structure of DeepONet, in which we have {\it branch} and {\it truck} networks. To accommodate two separate input functions, instead of feeding them into one branch network as we did in the  Fig.~\ref{fig:deeponet}, we create two separate {\it branch} networks: one for each input function. The two branch sub-networks will merge into one network via element-wise multiplication output vectors of each sub-network. 
Specifically, the enhanced DeepONet can be expressed as
\begin{strip}
\begin{align}
\label{eq:edeeponet_equ}
    \left|G(u)(y)-\langle \underbrace{\mathbf{g}\left(u\left(x_{1}\right), u\left(x_{2}\right), \cdots, u\left(x_{m}\right)\right)\odot\mathbf{h}\left(v\left(x_{1}\right), v\left(x_{2}\right), \cdots, v\left(x_{n}\right)\right)}_{\text {branch }}, \underbrace{\mathbf{f}(y)}_{\text {trunk }} \rangle \right|<\epsilon
\end{align}
\end{strip}
where operation $\odot$ indicates the element wise multiplication of two vectors.

\begin{figure}[h]
    \centering
    \subfigure[]{\includegraphics[width=0.49\linewidth]{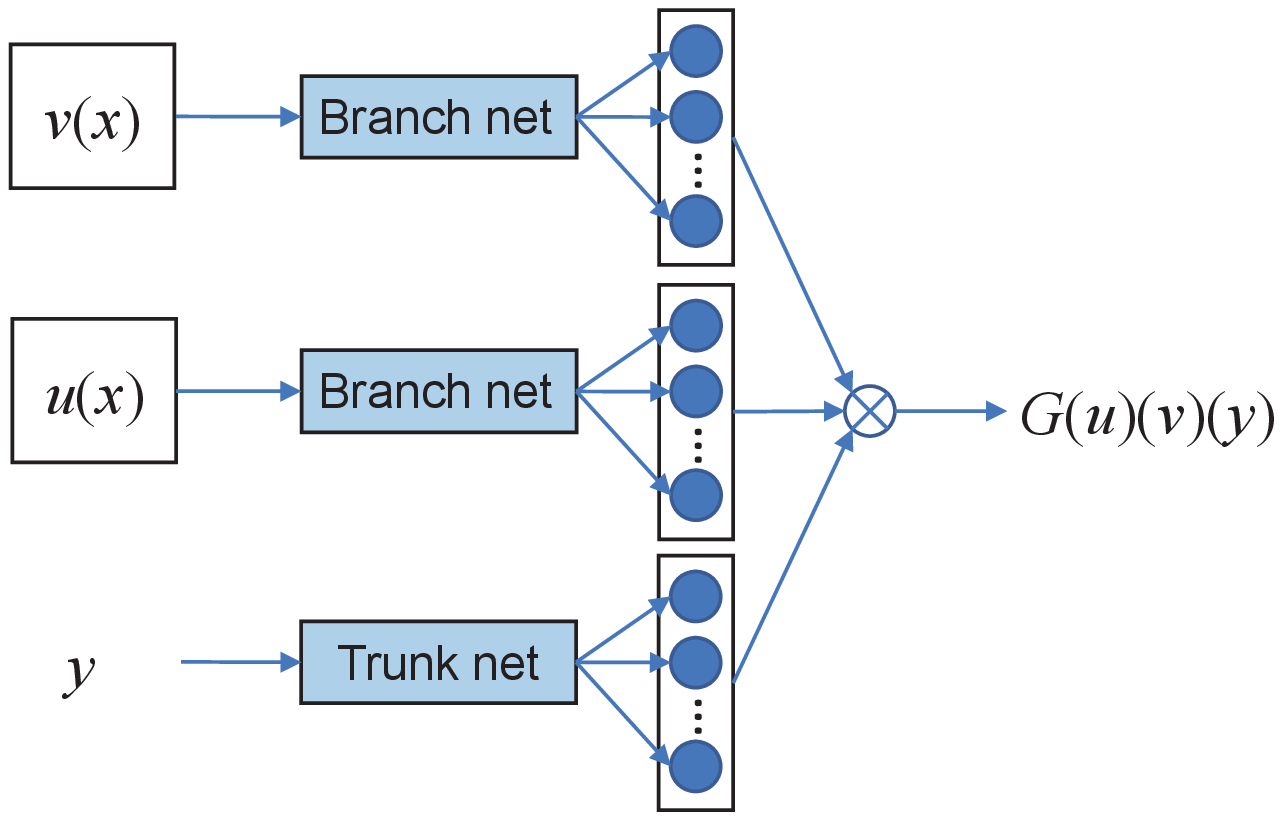}\label{fig:edeeponet1}}
    \subfigure[]{\includegraphics[width=0.49\linewidth]{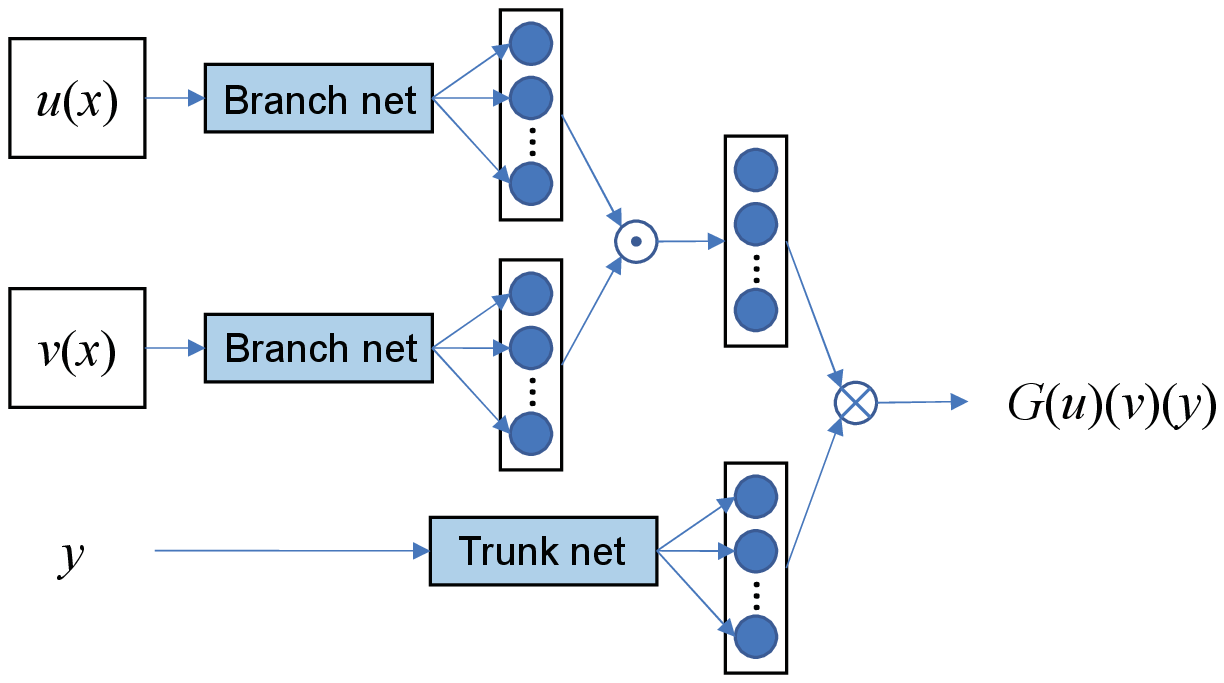}\label{fig:edeeponet2}}
    \caption{The proposed enhanced DeepONet, {\it EDeepONet} for universal  operator approximation for two input functions. (a) The first flat version  (b) The second hierarchical version.}
    \label{fig:edeeponet}
\end{figure}

Such element wise multiplication of two branch sub-network structures is inspired by the observation that the input and output networks are merged via inner product of their output vectors as shown in unstacked DeepONet structure shown in Fig.\ref{fig:deeponet1}. As a result, the new input function and their branch network should be connected with the other branch network and output truck network via inner product again as shown in Fig.~\ref{fig:edeeponet1}. Since all three sub-networks are in the same level, this version is called flat version of  {\it EDeepONet}. 

We can further re-organize the two branch sub-networks and merge them into one combined branch network via element wise multiplication first and then 
 After this we perform the inner product of the branch network vector with the truck network vector, which is the same operation for the DeepONet as shown in \eqref{eq:univ_op_approx}, to generate the final output for the whole neural network. The resulting architecture of EDeepONet can be shown in Fig.~\ref{fig:edeeponet2}. This version is called hierarchical version of {\it EDeepONet}. As we can see mathematically, the flat version and hierarchical version are equivalent, both of them can be represented by \eqref{eq:edeeponet_equ}. In our implementation, only the hierarchical one is implemented. 

Due to the limited space, we will not present the detailed mathematical proof of the proposed EDeepONet architecture for two or multiple input functions based on the universal operator approximation theorem framework~\cite{Chen:TNN'95}.  

\begin{figure}[!]
    \centering
    \includegraphics[width=0.55\linewidth]{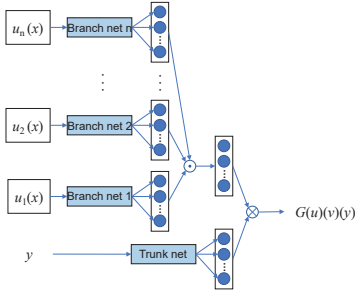}
    \caption{The proposed enhanced DeepONet, {\it EDeepONet} for universal  operator approximation for multi-function input.}
    \label{fig:edeeponetmulti}
\end{figure}

As we can see that the proposed EDeepONet can be trivially extended to consider more than two input functions as we just need to build more sub-branch networks for each input functions. Then we can perform the element wise multiplication on all the sub-branch networks to merge them into one branch vector, which then will perform inter production with the truck sub-network to produce the final output result as shown in Fig.~\ref{fig:edeeponetmulti}. 

\section{Experimental results and discussions}
\label{sec:results}
In this section, we present results of the proposed {\it EDeepONet} and  compare it with the baseline neural network and fully connected neural networks on two partial differential examples. We implement is based on the open-sourced DeepXDE package~\cite{DeepXDE_arxiv_2019,Lu:SIAM'21} and figures were drawn using OriginLab software. The ground truth results for comparison were obtained by the finite difference method implemented in DeepXDE package.  

Also for the fair comparison, all the networks we implemented will have similar number of parameters or weights. The training are set up so that the errors for training and test are similar over 1000 epochs so that no obvious  overfitting and overfitting are observed. 

\subsection{Numerical results on the diffusion equation}
The first example is a diffusion partial differential equation as shown  \eqref{eq:diffusion}. In physics, it describes  the macroscopic behavior of many micro-particles in Brownian motion, resulting from the random movements and collisions of the particles. For instance, heat transfer behavior in solid can be represented by the diffusion PDE. For instance $s(x,t)$ represent the temperature at location $x$ and time $t$ of a metal wire. 

Mathematically, \eqref{eq:diffusion} describes a diffusion system  
with periodic boundary condition and initial condition $s(x,0)$ = $v(x)$, where $a(x)=0.1+0.1\times\frac{u(x)+u(1-x)}{2}$ is a location  dependent  diffusion coefficient. $u(x)$ and $v(x)$ are random functions over location variable $x$. 
\begin{equation}
\label{eq:diffusion}
    \frac{\partial s(x,t)}{\partial t}-a(x) \frac{\partial^{2} s(x,t)}{\partial x^{2}}=0, \quad x \in(0,1), t \in(0,1]
\end{equation}
\begin{figure}[h]
    \centering
    \subfigure[]{\includegraphics[width=0.45\linewidth]{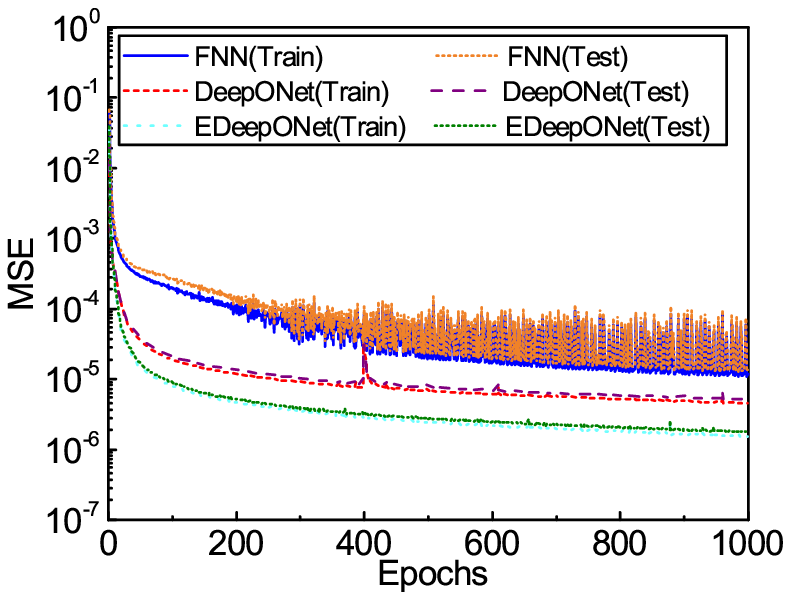}\label{fig:loss2}}
    \subfigure[]{\includegraphics[width=0.45\linewidth]{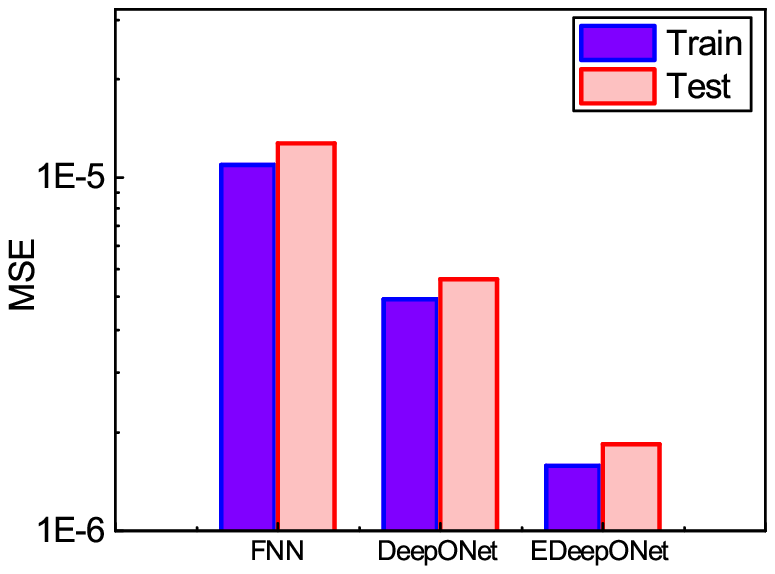}\label{fig:error2}}
    \caption{(a) The MSE of training set and test set for each epochs for the diffusion example. (b) The best training and test MSE for FNN, DeepONet and EDeepONet.}
\end{figure}

To demonstrate three methods, we set learning rate and batch size to 0.0001 and 10000, respectively. The training set contains $10^7$ samples and the test set has $10^5$ samples. 

The mean square errors (MSE) for each epochs are plotted in Fig.~\ref{fig:loss2}. Compared with FNN, DeepONet and EDeepONet have faster convergence speed, especially at the beginning of the training, which means that they can achieve same accuracy with fewer epochs -- an important advantage of the DeepONet over existing DNN networks. Actually this is consistent with the observation of DeepONet performance reported in~\cite{Lu:NMI'21} compared to other existing deep neural networks. 

Table~\ref{tab:diff_error_results} shows the comparison of the best MSE of training and test results for the three methods. Column 4 and 5 also show the accuracy improvement of EDeepONet over the baseline methods. Fig.~\ref{fig:error2} intuitively shows the best training and test MSE for FNN, DeepONet, and EDeepONet.

\begin{table}[h]
\center 
\label{tab:diff_error_results}
\caption{The best training and test MSE comparison for FNN, DeepONet and EDeepONet for the diffusion example}
\begin{tabular}{c | c | c | c | c }
\hline \hline 
Networks & \makecell{Training MSE} & Test MSE  &  \makecell{Training \\ Improv.} &  \makecell{Test \\ Improv.}  \\ \hline 
FNN & $ 1.08 \times 10^{-5}$ & $1.25 \times 10^{-5}$ &  7.05X & 7.10X \\
DeepONet & $4.53 \times 10^{-6}$ & $5.14 \times 10^{-6}$ & 2.96X & 2.92X \\
EDeepONet & $ 1.53 \times 10^{-6}$  & $ 1.76 \times 10^{-6}$ & -- & --  \\
\hline \hline
\end{tabular}
\end{table}

As we can see, among them, the MSE of FNN is worst and EDeepONet has the best MSE among them. DeepONet still has advantage over FNN, which has already been illustrated in~\cite{Lu:NMI'21}. Overall, the EDeepONet is  about $7X$  or about {\bf  order of magnitude more accurate}  than FNN and about $3X$ more accurate than DeepONet, which is quite significant.   The results demonstrate that our proposed method has a big improvement compared with FNN and DeeONet.

\begin{figure}[!]
    \centering
    \subfigure[]{\includegraphics[width=0.35\linewidth]{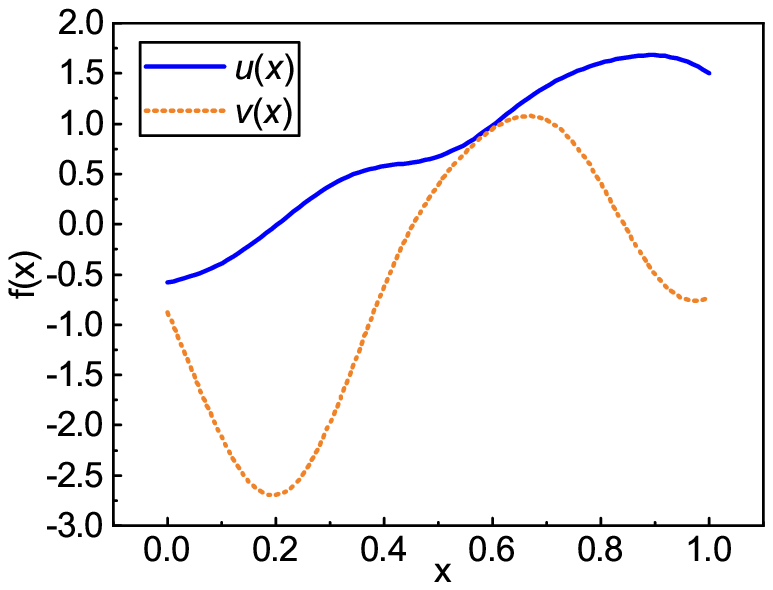}\label{fig:fnnuv2}}
    \subfigure[]{\includegraphics[width=0.35\linewidth]{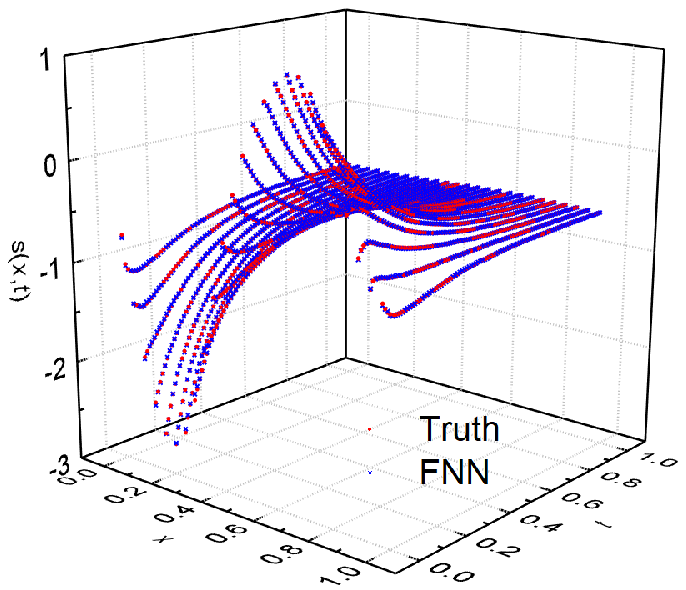}\label{fig:fnns2}}
    \subfigure[]{\includegraphics[width=0.35\linewidth]{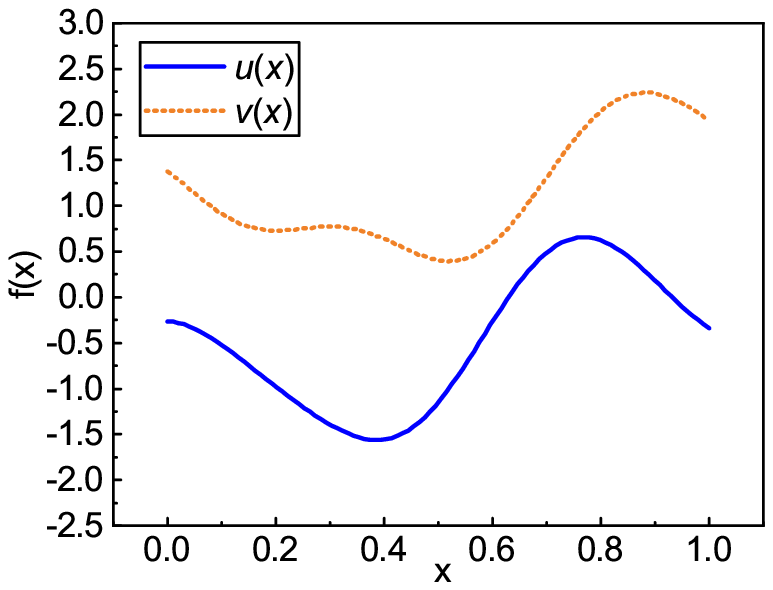}\label{fig:deeponetuv2}}
    \subfigure[]{\includegraphics[width=0.35\linewidth]{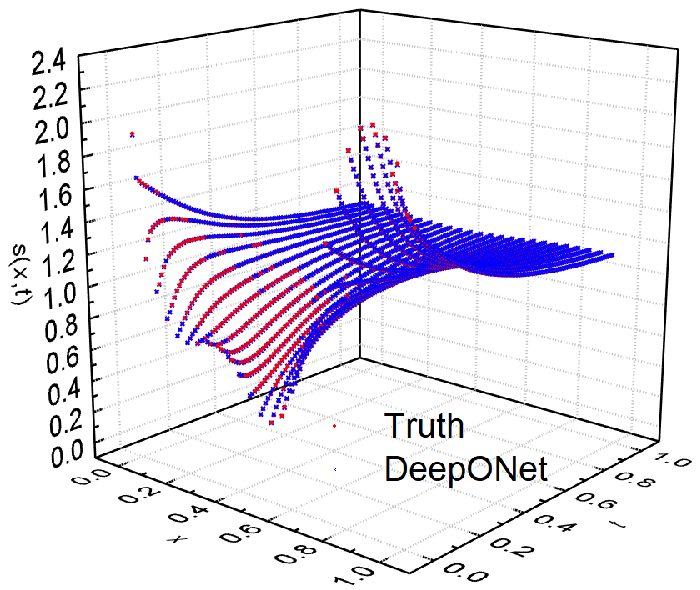}\label{fig:deeponets2}}
    \subfigure[]{\includegraphics[width=0.35\linewidth]{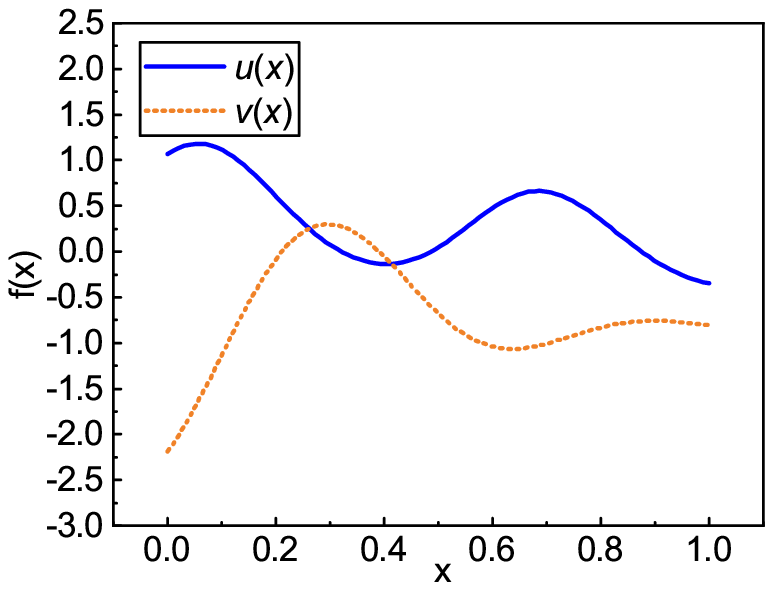}\label{fig:edeeponetuv2}}
    \subfigure[]{\includegraphics[width=0.35\linewidth]{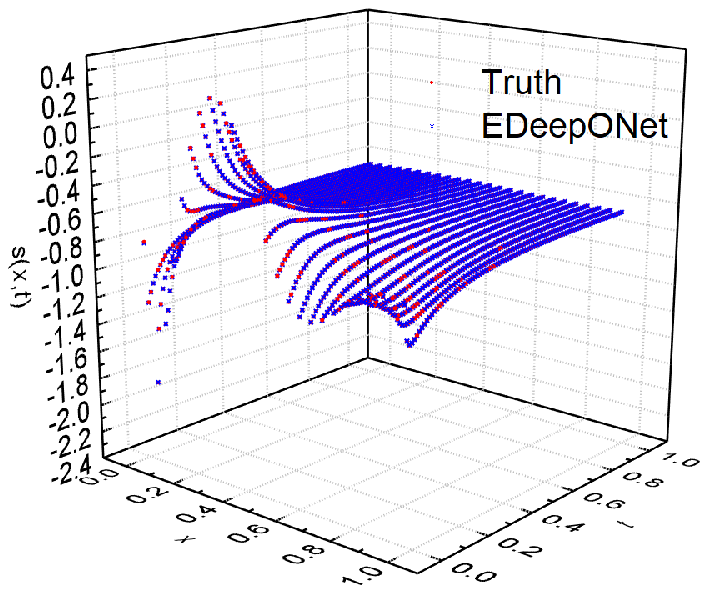}\label{fig:edeeponets2}}
    \caption{(a)(c)(e) The input functions $u(x)$,$v(x)$ and (b)(d)(f) the solution $s(x, t)$ for (a)(b) FNN, (c)(d) DeepONet, and (e)(f) EDeepONet in the diffusion equation.}
    \label{fig:diff}
\end{figure}

Fig.~\ref{fig:diff} shows the solution of PDE, which is obtained by FNN, DeepONet, and EDeepONet. We employ the finite difference numerical method to capture the
ground truth. Then, we compare the predictions by neural networks with ground truth. We show the two input functions $u(x)$ and $v(x)$ in Fig.~\ref{fig:fnnuv2}, Fig.~\ref{fig:deeponetuv2}, and Fig.~\ref{fig:edeeponetuv2}, which are generated by random functions over $x$. As a result, for three networks, their input functions are different as shown in Fig.~\ref{fig:diff}.

Clearly, Fig.~\ref{fig:fnns2} shows that there are some clear mismatches between FNN and ground truth. This is also the case for the DeepONet case shown in Fig.~\ref{fig:deeponets2}. For results from the EDeepONet case, we can see from  Fig.~\ref{fig:edeeponets2}  that the prediction and truth agree well with each other.

\begin{figure}[!]
    \centering
    \subfigure[]{\includegraphics[width=0.4\linewidth]{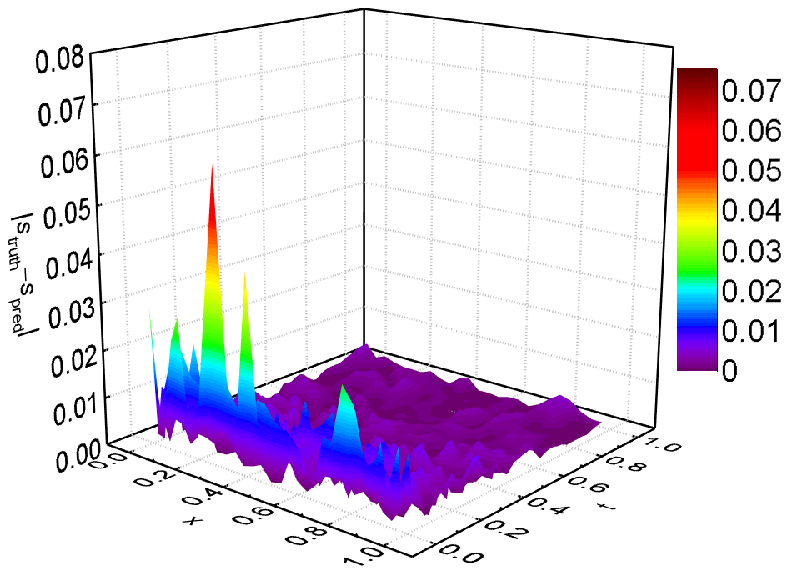}\label{fig:fnne2}}
    \subfigure[]{\includegraphics[width=0.4\linewidth]{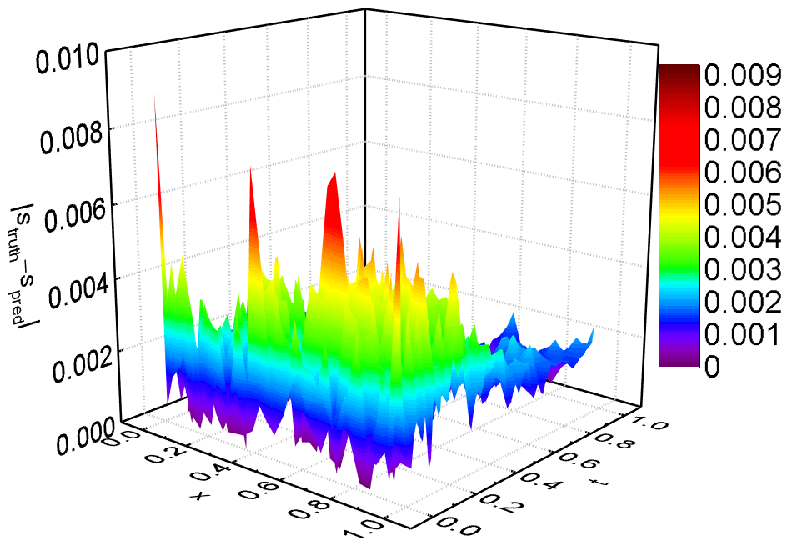}\label{fig:deeponete2}}
    \subfigure[]{\includegraphics[width=0.4\linewidth]{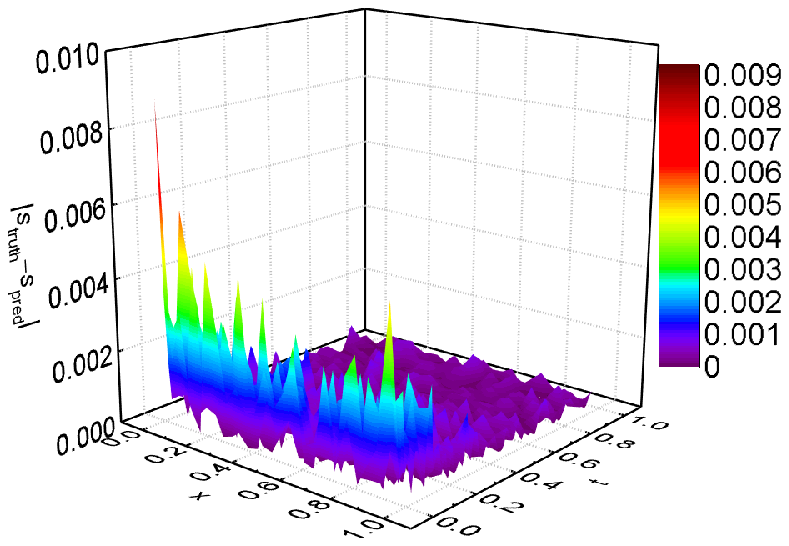}\label{fig:edeeponete2}}
    \caption{The absolute error $|s_{\text{pred}}-s_{\text{truth}}|$ for (a) FNN, (b) DeepONet, and (c) EDeepONet in the diffusion equation.}
    \label{fig:diff_error}
\end{figure}

To further precisely describe the difference between the prediction and ground truth, we plot absolute error $|s_{\text{pred}}-s_{\text{truth}}|$ in Fig.~\ref{fig:diff_error}. As we can see, the maximum error of EDeepONet is smaller than those of FNN and DeepONet. As we can see that EDeepONet achieves close to order of magnitude maximum error reduction compared to FNN. Therefore, our proposed EDeepONet has better performance than FNN and DeepONet.

\subsection{Numerical results on advection and diffusion equation}
To further demonstrate the strength of the proposed EDeepONet, the second example is the advection (or convection) and diffusion equation example shown in \eqref{eq:adv_diff} with periodic boundary condition and initial condition $s(x,0) =v(x)$, where $a(x)= 1+0.1\times\frac{u(x)+u(1-x)}{2}$ a location  dependent  diffusion coefficient. Since we have the $\frac{\partial s}{\partial x}$ term in \eqref{eq:adv_diff}, which represents the advection or convection effect for the variable $s$.  Advection effect is the transport of a substance or quantity by bulk motion of a fluid. For thermal systems, it typically called convection for the air flow induced convection. 
This is the same example used in the DeepONet work also~\cite{Lu:NMI'21}. 
\begin{equation}
\label{eq:adv_diff}
    \frac{\partial s}{\partial t}+\frac{\partial s}{\partial x}-a(x) \frac{\partial^{2} s}{\partial x^{2}}=0, \quad x \in(0,1), t \in(0,1]
\end{equation}

\begin{figure}[!ht]
    \centering
    \subfigure[]{\includegraphics[width=0.45\linewidth]{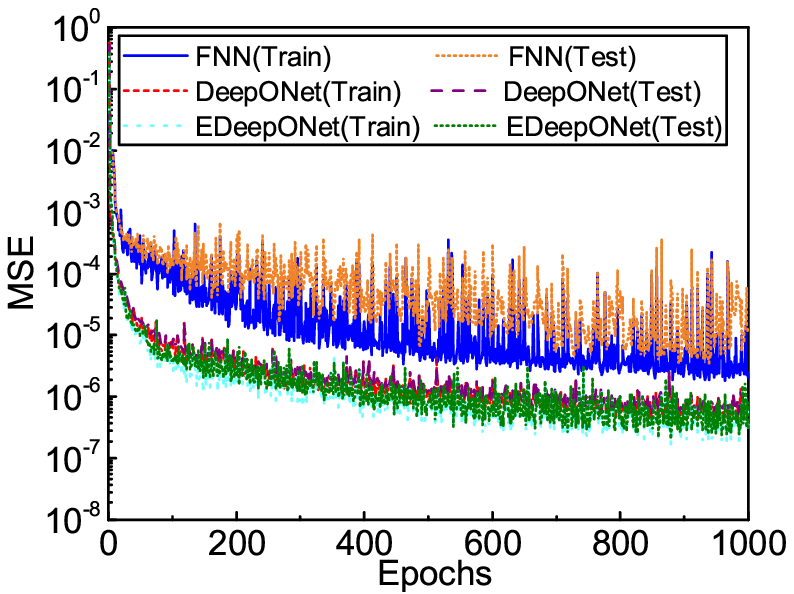}\label{fig:loss}}
    \subfigure[]{\includegraphics[width=0.45\linewidth]{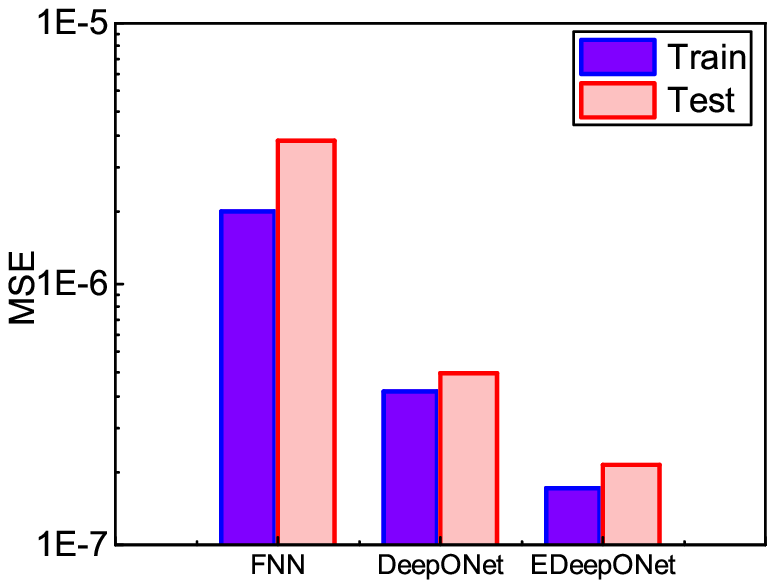}\label{fig:error}}
    \caption{(a) The MSE of training set and test set for each epochs for the advection-diffusion example. (b) The best training MSE for FNN, DeepONet, and EDeepONet.}
\end{figure}

Fig.~\ref{fig:loss} shows the mean square error (MSE) history of training and test set. Similarly, DeepONet and EDeepONet have faster convergence speed than FNN.

Table~\ref{tab:diff_advection_error_results}  shows the comparison of  best MSE of training and test results for the three methods for the advection-diffusion example. Column 4 and 5 also show the accuracy improvement of EDeepONet over the baseline methods. Again Fig.~\ref{fig:error} shows the same best training and test MSE for FNN, DeepONet, and EDeepONet.

\begin{table}[h]
\center 
\label{tab:diff_advection_error_results}
\caption{The best training and test MSE comparison for FNN, DeepONet and EDeepONet for the advection-diffusion example}
\begin{tabular}{c | c | c | c | c}
\hline \hline 
Network & Training MSE  & Test MSE & \makecell{Training \\Improv.} & \makecell{Test \\Improv.} \\ \hline 
FNN & $ 1.90 \times 10^{-6}$ & $3.54 \times 10^{-6}$ & 11.5X & 17.43X \\
DeepONet & $3.87 \times 10^{-7}$ & $4.55 \times 10^{-7}$ & 2.35X & 2.24X \\
EDeepONet & $ 1.65 \times 10^{-7}$  & $ 2.03 \times 10^{-7}$ & -- & -- \\
\hline \hline
\end{tabular}
\end{table}

As we can see that in the advection-diffusion case, EDeepONet is $11.5X$ more accurate for training and $17.43X$  more accurate than FNN, which is very impressive. However, for the comparison against DeepONet, the EDeepONet is about $2X$ more accurate. This further demonstrates the advantage of the proposed EDeepONet networks over existing state of the art methods. 

\begin{figure}[!]
    \centering
    \subfigure[]{\includegraphics[width=0.4\linewidth]{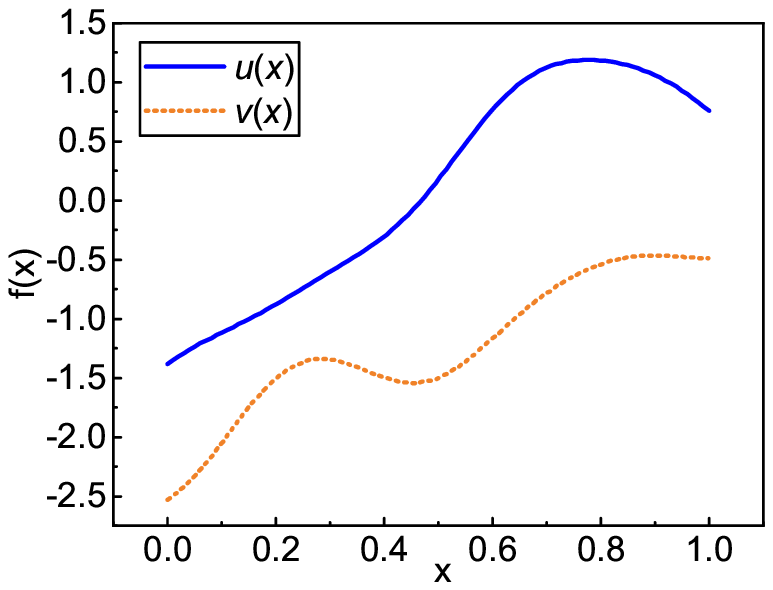}\label{fig:fnnuv}}
    \subfigure[]{\includegraphics[width=0.4\linewidth]{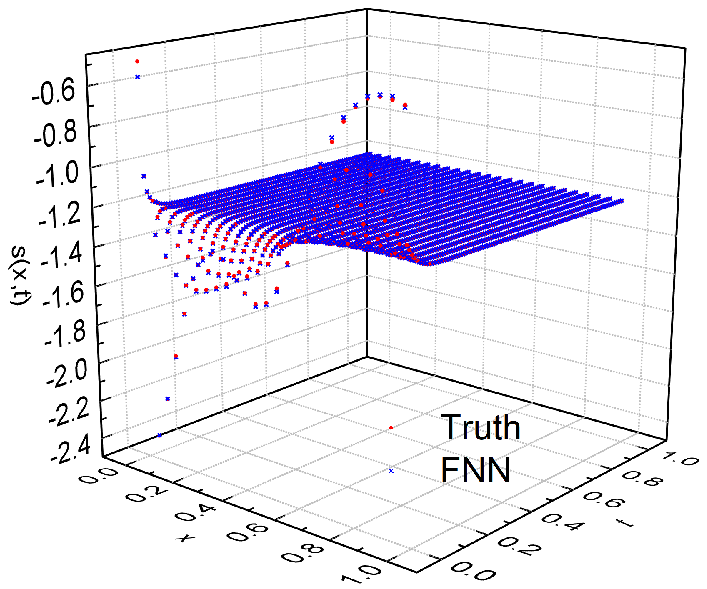}\label{fig:fnns}}
    \subfigure[]{\includegraphics[width=0.4\linewidth]{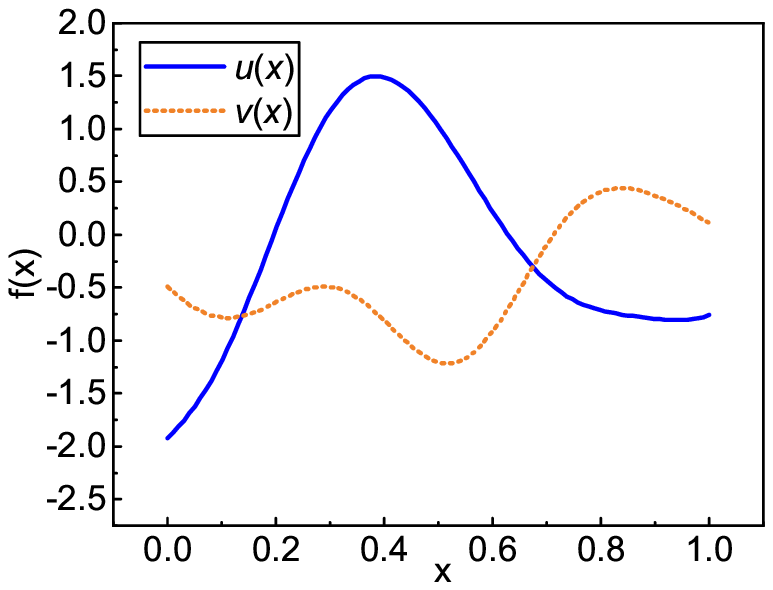}\label{fig:deeponetuv}}
    \subfigure[]{\includegraphics[width=0.4\linewidth]{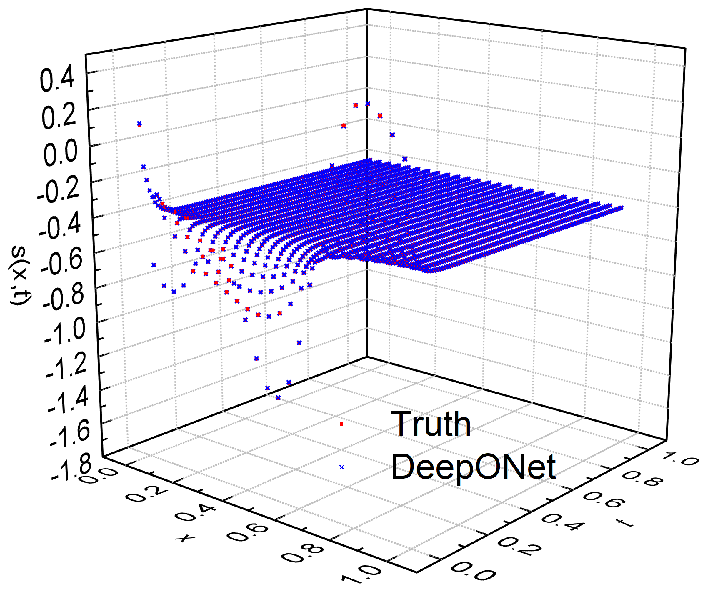}\label{fig:deeponets}}
    \subfigure[]{\includegraphics[width=0.4\linewidth]{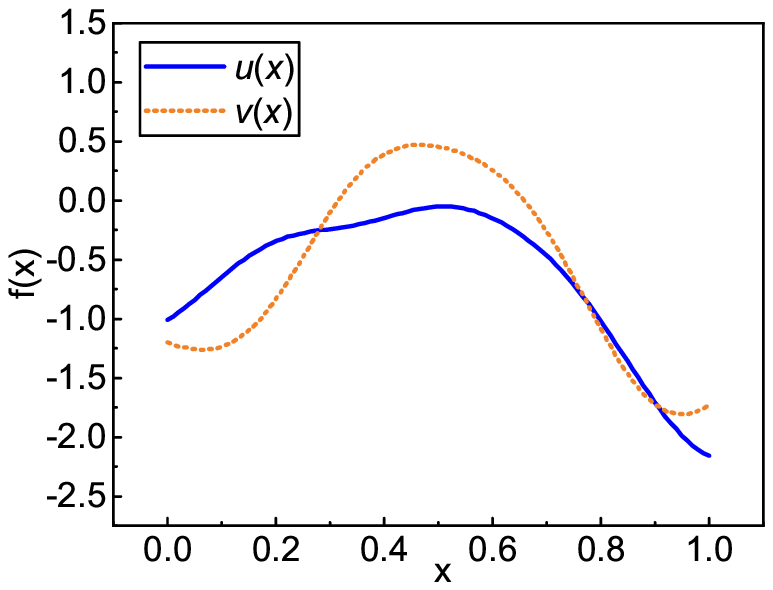}\label{fig:edeeponetuv}}
    \subfigure[]{\includegraphics[width=0.4\linewidth]{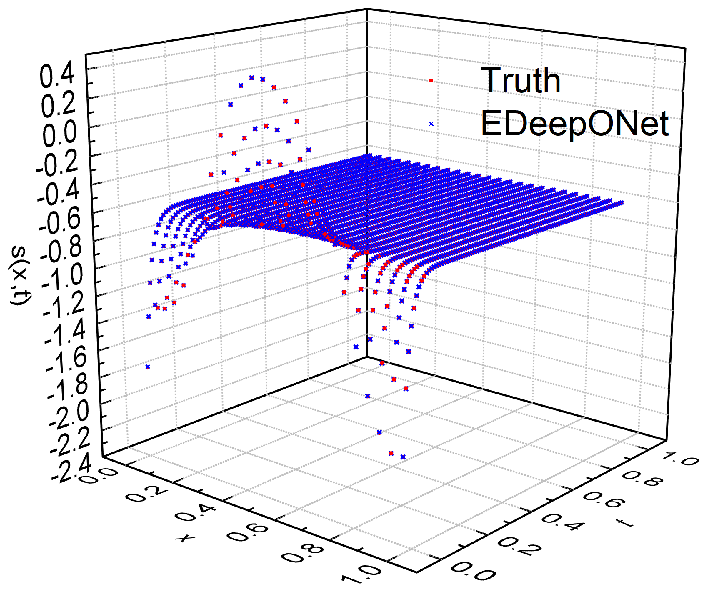}\label{fig:edeeponets}}
    \caption{(a)(c)(e) The input functions $u(x)$,$v(x)$ and (b)(d)(f) the solution $s(x, t)$ for (a)(b) FNN, (c)(d) DeepONet, and (e)(f) EDeepONet in the advection and diffusion equation.}
    \label{fig:conv_diff}
\end{figure}

Again for the second example, we also 
show the solution of PDE in Fig.~\ref{fig:conv_diff} for FNN, DeepONet, and EDeepONet. As we can see that both FNN and DeepONet results show some visible difference compared to the group truth. 
However, the prediction by EDeepONet match better with  the truth, as show in Fig.~\ref{fig:edeeponets}.

\begin{figure}[!]
    \centering
    \subfigure[]{\includegraphics[width=0.4\linewidth]{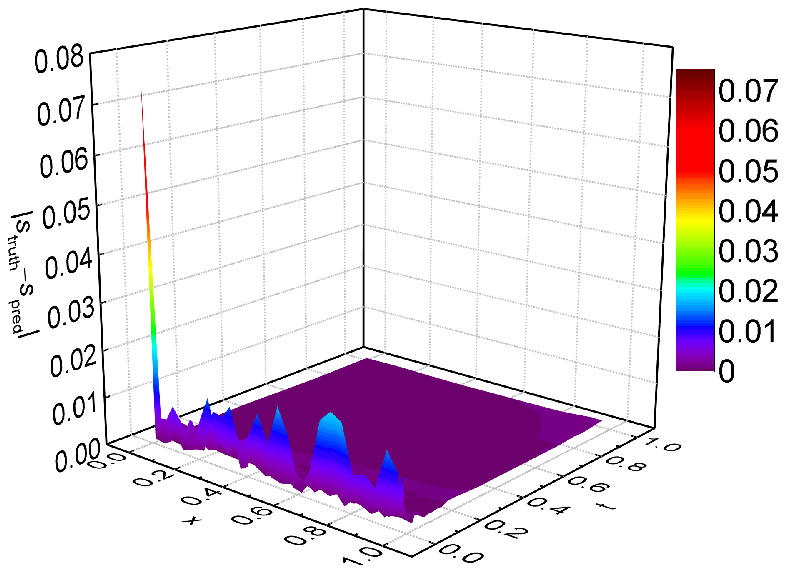}\label{fig:fnne}}
    \subfigure[]{\includegraphics[width=0.4\linewidth]{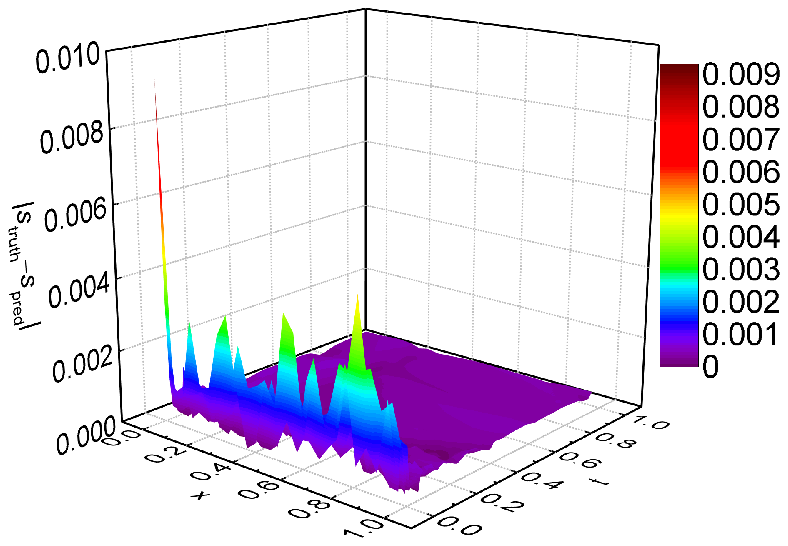}\label{fig:deeponete}}
    \subfigure[]{\includegraphics[width=0.4\linewidth]{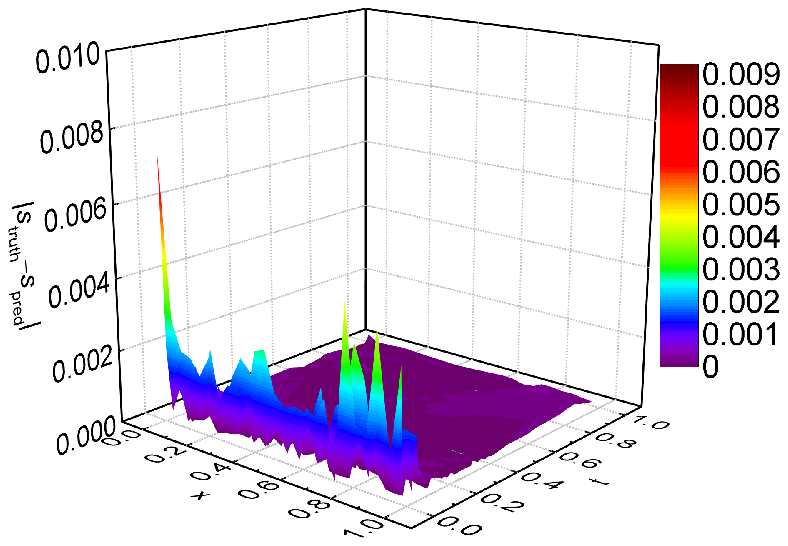}\label{fig:edeeponete}}
    \caption{The absolute error $|s_{\text{pred}}-s_{\text{truth}}|$ for (a) FNN, (b) DeepONet, and (c) EDeepONet in the advection and diffusion equation.}
    \label{fig:conv_diff_error}
\end{figure}

Fig.~\ref{fig:conv_diff_error} shows absolute error $|s_{\text{pred}}-s_{\text{truth}}|$ between the prediction and truth. As we can see, the maximum error of EDeepONet is still smaller than those of FNN and DeepONet, which is consistent with the diffusion example.

\section{Conclusion and future direction}
\label{sec:concl}
In this work,  we have proposed new Enhanced DeepONet or {\it EDeepONet} high level neural network structure, in which two input functions are represented by two branch DNN sub-networks, which are then connected with output truck network via inner product to generate the output of the whole neural network.  The proposed {\it EDeepONet} structure can be easily extended to deal with multiple input functions. Our numerical results on modeling two partial differential equation examples have shown that the proposed enhanced DeepONet is about  $7X$ to $17X$ or {\bf  one order of magnitude}  more accurate than the fully connected neural network and is about $2X$-$3X$ more accurate than the simple extended DeepONet  for both training and test. 

In the future, we will consider a few research directions. First we  will explore  more partial differential equation examples and perform the comparisons with fully-connected neural networks and DeepONet baseline method. Second, we   study the operator modeling for partial differential equations for multiple input functions (at least three or more) to validate the proposed method for considering  general multiple input functions.

\bibliographystyle{IEEEtran}
\bibliography{refer}

\end{document}